\documentclass{article}

\usepackage[final]{timeseries_workshop}

\usepackage[utf8]{inputenc} 
\usepackage[T1]{fontenc}    
\usepackage{hyperref}       
\usepackage{url}            
\usepackage{booktabs}       
\usepackage{amsfonts}       
\usepackage{nicefrac}       
\usepackage{microtype}      
\usepackage{xcolor}         
\usepackage{tabularx}
\usepackage{threeparttable}
\usepackage{graphicx} 
\usepackage{graphicx}
\usepackage{subcaption}
\title{Unveiling the Potential of Text in High-Dimensional Time Series Forecasting}

\author{
  Xin Zhou\thanks{xin.zhou@monash.edu}, Weiqing Wang, Shilin Qu, \\
  Monash University, \\
  Melbourne, Victoria, Australia \\
  \and
  Zhiqiang Zhang, \\
  Southeast University, \\
  Nanjing, China \\
  \and
  Christoph Bergmeir, \\
  University of Granada, \\
  Granada, Spain \\
}
 
\begin{document}

\maketitle

\begin{abstract}
Time series forecasting has traditionally focused on univariate and multivariate numerical data, often overlooking the benefits of incorporating multimodal information, particularly textual data. In this paper, we propose a novel framework that integrates time series models with Large Language Models to improve high-dimensional time series forecasting.
Inspired by multimodal models, our method combines time series and textual data in the dual-tower structure. This fusion of information creates a comprehensive representation, which is then processed through a linear layer to generate the final forecast. Extensive experiments demonstrate that incorporating text enhances high-dimensional time series forecasting performance. This work paves the way for further research in multimodal time series forecasting.
\end{abstract}

\section{Introduction}

Time series forecasting plays a vital role in various domains such as finance \cite{kdd_financial, ICDEstock, ICDEfinancial}, traffic \cite{tslanet,ICDEtraffic1, ctfnet}, healthcare \cite{nipsconvid, ICDEcovid, ICDEhealth}, and weather prediction \cite{convtimenet,nipsconformal, ICDEweather}. 
Traditional deep learning-based methods focus on time series analysis, relying purely on the historical numerical values of individual series.
However, they often overlook the potential of textual data, particularly in high-dimensional time series forecasting, which involves a large number of channels and complex, diverse temporal patterns\cite{highdim, fnets}. In such a scenario, textual information can provide valuable supplementary insights. 

The recent success of Large Language Models in Natural Language Processing and Computer Vision has sparked interest in using textual data to enhance time series forecasting~\cite{XI2023103254,Jia_Wang_Zheng_Cao_Liu_2024}, however, their data are low-dimensional.
Thus, incorporating textual information into high-dimensional time series forecasting remains unexplored. 
Our work aims to bridge this gap by being the first to integrate textual data into high-dimensional time series forecasting, introducing a novel challenge and paving the way for future research in this field.
We propose that integrating numerical time series data with textual information allows models to develop a more comprehensive understanding of underlying dynamics, ultimately improving forecasting accuracy.

In our work, we propose a general framework - TextFusionHTS, integrating textual information into high-dimensional time series forecasting, focusing on the challenges of extracting and capably fusing textual data with time series inputs. 
Our key contributions are as follows:

(1) Framework for Text-Integrated High-Dimensional Time Series Forecasting: We introduce a new framework TextFusionHTS that integrates textual data into high-dimensional time series analysis. By integrating text-based information, we open up new possibilities for high-dimensional time series forecasting and pave the way for further research in multimodal time series analysis;

(2) Experiments Validation: We conduct extensive experiments demonstrating the effectiveness of incorporating textual data into high-dimensional time series forecasting. Our results~\footnote{\href{https://github.com/xinzzzhou/TextFusionHTS.git}{https://github.com/xinzzzhou/TextFusionHTS.git}} reveal that textual information can notably enhance predictive performance, highlighting the untapped potential of multimodal inputs in this field.
\section{Related Works}

\subsection{Time Series Forecasting}
Time series forecasting methods have employed various deep learning techniques, including Linear \cite{timemixer,dlinear, tsmixer}, Convolutional Neural Networks \cite{convtimenet, ctfnet, moderntcn}, and Transformer \cite{fredformer, gpht, nipsenercy, pathformer}. Most of them mainly solve the challenge of capturing short-term and long-term dependencies. PatchTST \cite{patchtst} has proven to be a leading technique that divides the series window into patches, allowing it to capture the complex global and local relationships within individual series. Recently, LLM-based methods, e.g., TimeLLM \cite{timellm}, TEMPO \cite{tempo}, TEST \cite{test}, and TimeCMA~\cite{liu2024timecmallmempoweredtimeseries} have explored adapting LLMs for time series tasks by converting time series data into text formats that LLMs can effectively process.

\subsection{Multimodal Time Series Forecasting}
Several methods have explored integrating different data types. \cite{isa24} introduced a framework that incorporates video information into time series forecasting, while Multimodal Time-series Analysis~\cite{XI2023103254} fuses textual, visual, and audio data to forecast gifting behavior on live-streaming platforms. GPT4MTS~\cite{Jia_Wang_Zheng_Cao_Liu_2024} proposes a news multimodal dataset based on GDELT and provides a pipeline for data extraction and processing.
These efforts demonstrate the potential of multimodal forecasting, though they often encounter challenges due to the lack of published code or data.
Additionally, MST-GAT~\cite{DING2023527} utilizes graph attention mechanisms to manage multimodal inputs, specifically designed for anomaly detection in time series data. 
While these methods show promise in integrating textual data, the performance of fusing different modalities in high-dimensional time series forecasting remains unclear.

\section{Method}

\begin{figure}[htbp]
    \centering
    \includegraphics[width=0.6\linewidth]{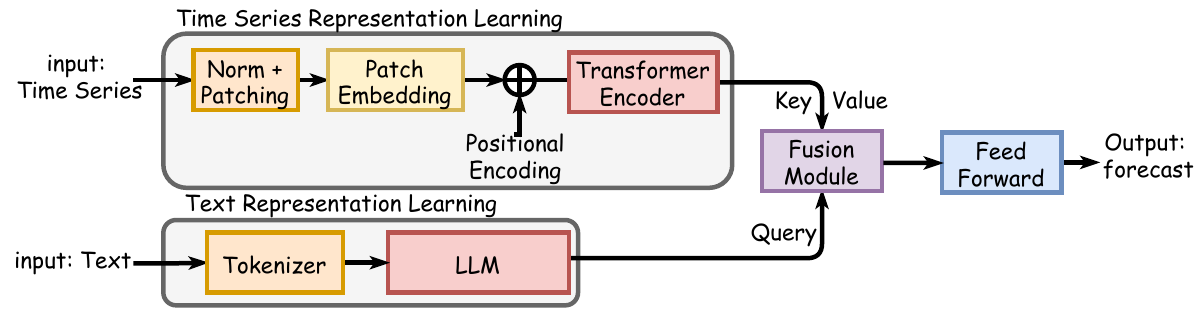}
    \caption{Dual-tower structure of TextFusionHTS.}
    \label{fig:pipeline}
\end{figure}

Inspired by the dual-tower architecture commonly used in multimodal methods~\cite{koh2024generating,sun2024generative}, we propose a framework TextFusionHTS that integrates Large Language Models with time series models, as shown in Figure~\ref{fig:pipeline}.

\subsection{Time Series Representation Learning} 
We first use a patch-based method, PatchTST, to extract the representation of the time series data. 
The input series $x \in \mathbb{N}^{l}$ with $l$ length is divided into $p$ patches.
Using the patch-based method is crucial in dynamically using static textual data.
Following PatchTST, dividing time series into patches allows the model to capture local temporal patterns within each patch. Since textual data is static, it's important to assign textual data to different interactions with various time series patches of the series. For example, the text may provide context that is more relevant to certain periods or patterns in the series. Therefore, using a patch-based method ensures that the model can apply the static text data in a context-sensitive manner for different time series windows.

PatchTST generates embedding for each series as follows, 
\begin{equation}
    z_{ts} = PatchTST(x) \in \mathbb{R}^{p \times d_{ts}}, \quad K = V = z_{ts}
\end{equation}
where $d_{ts}$ denotes the dimension of each patch’s representation. This matrix serves as the key ($K$) and values ($V$) for the subsequent attention mechanism for the fusion step.

\subsection{Text Representation Learning}
To incorporate textual data, we use Meta Llama-3.1 8B, the state-of-the-art open-source LLM, to process the text description corresponding to the time series. 
The text is tokenized and converted into a set of toke embeddings. Let the set of token embeddings be $\{t_i \in \mathtt{R}^{d_{tx}} | i=1,2,...,n \}$ where $d_{tx}$ is the dimension of text representation, n is the number of tokens. 
We compute the average of all token embeddings to get full semantic meaning of text,
\begin{equation}
    z_{tx} = \frac{1}{n} \sum_{i=1}^{n} t_i \in \mathbb{R}^{d_{tx}}, \quad Q = z_{tx}
\end{equation}
This averaged embedding $z_{tx}$ serves as the query ($Q$) in the cross-attention mechanism.

\subsection{Fusion Module}
We apply a cross-attention mechanism where the time series representations ($K$ and $V$) are combined with the textual representation ($Q$). This operation results in a fused representation $z = CrossAttention(Q, K, V) \in \mathbb{R}^{d}$ that integrates information from both modalities.

Finally, we use a Feed Forward transformation to map the fused representation $z$ to the forecast values $y \in \mathbb{R}^{h}$, resulting in the final predictions in $h$ horizon.

\section{Experiments}

\subsection{Experimental Settings}
We collect two high-dimensional time series forecasting datasets with textural data. \textit{Wiki-People}
records daily traffic of Wikipedia webpages from July 2015 to December 2016.
It was published in the Kaggle competition. We keep 3,857 full series published in the English language in the Wikipedia agent and crawl text data from the corresponding pages.
\textit{News}~\cite{moniz2018multisourcesocialfeedbackonline} is originally a large dataset of news items and their respective social feedback on Facebook. 
The collected data relates to a period of 8 months, between November 2015 and July 2016, accounting for 26,612 news items on the `Obama' topic. The time interval is 20 minutes.
Given that the time interval is 1 day for Wiki-People and 20 minutes for News, we set the input lengths to 7 and 9, corresponding to 1 week and 3 hours, respectively. 
The forecasting horizon $h$ is set to $\{7, 14, 21, 28, 35\}$ for Wiki-People and $ \{1, 3, 9, 12, 15\}$ for News.

All implementations are done in PyTorch and trained on a single NVIDIA A100 GPU. To ensure model convergence during training, we set the number of training epochs to 100 and employ an early stopping mechanism. Specifically, if the change in validation loss is less than $1 \times 10^{-4}$, the training stops, and testing begins.

\subsection{Result Comparison}
To assess the impact of incorporating textual information with our proposed framework, we conduct the following experiments for comparison: 1) using purely time series data as input, with PatchTST as the time series model; 2) using both time series data and textual data as input, with the framework introduced in Section 3.

\begin{table*}[ht]
 \centering
 \renewcommand{\arraystretch}{0.6}
 \small
 
\resizebox{\textwidth}{!}{%
 \begin{threeparttable}
    \caption{Experimental results. $TextFusionHTS_{wo}$ is using purely time series as input, and $TextFusionHTS_{wt}$ is using both time series and text as input. The best results are in \textbf{bold}.}
    \label{tab:main_results}
    \begin{tabular}{c|c|c|c|c|c|c|c|c|c|c}
        \hline
        Dataset & \multicolumn{10}{c}{News} \\
        \hline
         Metric & \multicolumn{5}{c|}{MAE} & \multicolumn{5}{c}{WAPE} \\
        \hline
        Horizon & 1 & 3 & 9 & 12 & 15 & 1 & 3 & 9 & 12 & 15 \\
        \hline
        $TextFusionHTS_{wo}$ & 0.3464 & 0.6072 & 0.8761 & 1.0431 & 1.2275 & 0.0625 & 0.1125 & 0.1695 & 0.2029 & 0.2390 \\
        $TextFusionHTS_{wt}$ & \textbf{0.3367} & \textbf{0.6040} & \textbf{0.8700} & \textbf{1.0350} & \textbf{1.2031} & \textbf{0.0607} & \textbf{0.1119} & \textbf{0.1683} & \textbf{0.2013} & \textbf{0.2343} \\
        \hline
        Dataset & \multicolumn{10}{c}{Wiki-People} \\
        \hline
         Metric & \multicolumn{5}{c|}{MAE} & \multicolumn{5}{c}{WAPE} \\
        \hline
        Horizon & 7 & 14 & 21 & 28 & 35 & 7 & 14 & 21 & 28 & 35 \\
        \hline
        $TextFusionHTS_{wo}$ & 0.5751 & 0.6509 & 0.7061 & 0.7506 & 0.7906 & 0.5209 & 0.5894 & \textbf{0.6382} & 0.6765 & 0.7108\\
        $TextFusionHTS_{wt}$ & \textbf{0.5734} & \textbf{0.6496} & \textbf{0.7053} & \textbf{0.7500} & \textbf{0.7873} & \textbf{0.5194} & \textbf{0.5883} & 0.6474 & \textbf{0.6759} & \textbf{0.7079}\\
        \hline
    \end{tabular}
 \end{threeparttable}}
\end{table*}

The Table~\ref{tab:main_results} summarizes the overall results across two datasets, showing that integrating textual data into our framework consistently improves performance in both Mean Absolute Error (MAE) and Weighted Absolute Percentage Error (WAPE). 
While the only exception, is that when the forecasting horizon $h$ is set to 21 on the Wiki-People dataset. 
When the time series data alone fails to provide adequate context, the additional textual data acts as a valuable complement, enhancing forecasting accuracy.

\subsection{Strategies for Text Feature Extraction}
To investigate the relation between text feature extraction and horizon. We compare three strategies for extracting textual features: using the [cls] token, the [bos] token, and the average of all token embeddings. Following are the results on the News dataset.

\begin{figure}[htbp]
    \centering
    \begin{subfigure}[b]{0.28\textwidth}
        \centering
        \includegraphics[width=\textwidth]{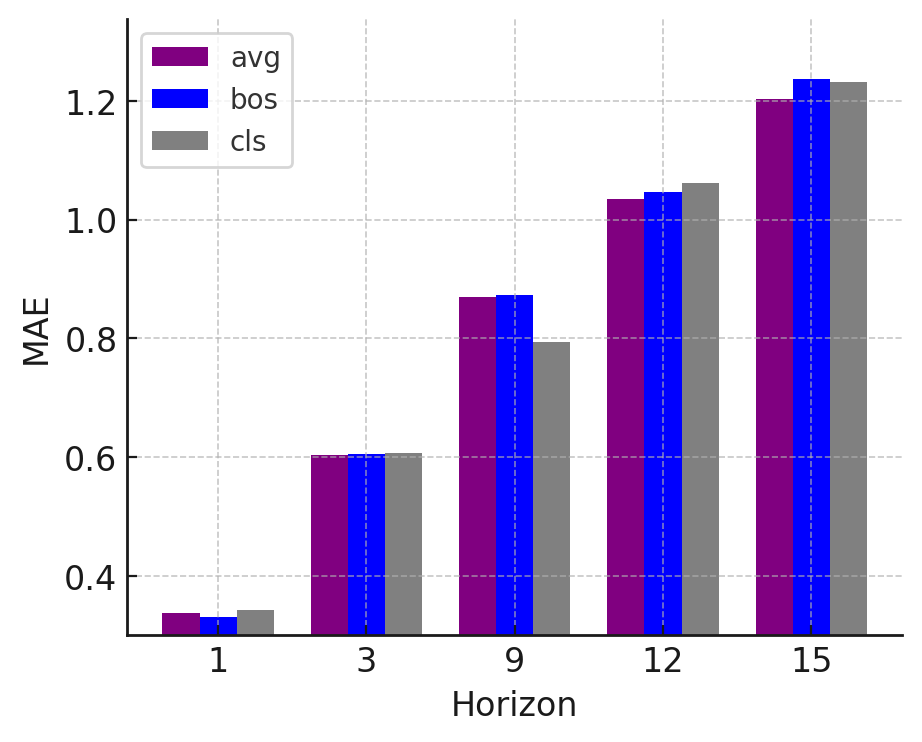}
        \caption{MAE}
        \label{fig:sub1}
    \end{subfigure}
    \hspace{0.5cm}
    \begin{subfigure}[b]{0.28\textwidth}
        \centering
        \includegraphics[width=\textwidth]{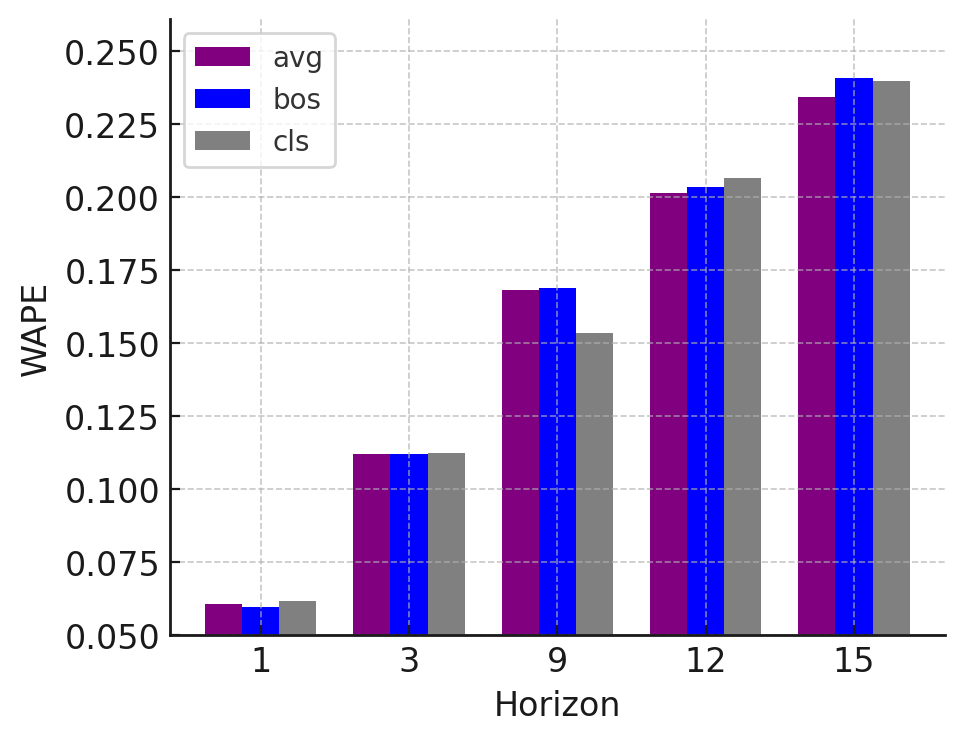}
        \caption{WAPE}
        \label{fig:sub2}
    \end{subfigure}
    \caption{Comparison of different strategies for text feature extraction. Here the average of all token embeddings, [bos] token embedding, and [cls] token embeddings are represented with purple, blue, and grey bars separately.}
    \label{fig:strategy}
\end{figure}

Figure~\ref{fig:strategy} shows that the choice of text feature extraction method significantly affects the model's performance. 
For shorter horizons, the differences between the various strategies are minimal. 
However, when the horizon extends to 9, using the [cls] token appears to underperform compared to other strategies, possibly because [cls] fails to capture the full semantic meaning of the entire sentence. 
For longer horizons, the [bos] token outperforms the others, indicating that it may contain more valuable contextual information.
\section{Conclusion}
In this paper, we investigated the potential of incorporating textual data into high-dimensional time series forecasting. 
Our experiments demonstrate that using textual data can notably enhance forecasting performance, particularly for longer horizons where time series data alone may fall short in providing sufficient context. This approach effectively addresses the limitations of traditional time series models, significantly boosting their predictive capabilities. These findings underscore the importance of multimodal approaches in time series forecasting. 
For future work, we plan to develop more advanced models capable of fusing additional modalities, such as images and audio, to further enhance forecasting accuracy. Additionally, we aim to conduct more comprehensive experiments on diverse datasets to fully explore the benefits of integrating multiple sources of information.

\section*{Acknowledgement}
This work contribution is part of the I+D+i project granted by C-ING-250-UGR23 co-funded by ``Consejería de Universidad, Investigación e Innovación'' and for the European Union related to the FEDER Andalucía Program 2021-2027''. This work is supported by the Knowledge Generation Project PID2023-149128NB-I00, funded by the Ministry of Science, Innovation and Universities of Spain. Christoph Bergmeir is supported by a María Zambrano Fellowship that is funded by the Spanish Ministry of Universities and Next Generation funds from the European Union.
Additionally, we gratefully acknowledge the support of Google for providing a travel grant, which enabled attendance at NeurIPS 2024. This funding has significantly contributed to our research efforts and facilitated essential academic exchange. We sincerely appreciate Google's generous support.

\bibliographystyle{apalike}
\bibliography{main}

\end{document}